\documentclass[review]{elsarticle}
\usepackage{amsmath}
\usepackage{amssymb}
\usepackage{lineno,hyperref}
\modulolinenumbers[5]
\usepackage[dvipsnames]{xcolor}
\usepackage{amsfonts}

\journal{Journal of \LaTeX\ Templates}

\begin{document}

\begin{frontmatter}

\title{SetMargin Loss applied to Deep Keystroke Biometrics \\ with Circle Packing Interpretation}

\author{Aythami Morales, Julian Fierrez, Alejandro Acien, Ruben Tolosana, \\ Ignacio Serna}
\address{School of Engineering, Universidad Autonoma de Madrid, Spain}
\ead{\{aythami.morales, julian.fierrez, alejandro.acien,  ruben.tolosana, ignacio.serna\}@uam.es}

\begin{abstract}
   This work presents a new deep learning approach for keystroke biometrics based on a novel Distance Metric Learning method (DML). DML maps input data into a learned representation space that reveals a ``semantic'' structure based on distances. In this work, we propose a novel DML method specifically designed to address the challenges associated to free-text keystroke identification where the classes used in learning and inference are disjoint. The proposed SetMargin Loss (SM-L) extends traditional DML approaches with a learning process guided by pairs of sets instead of pairs of samples, as done traditionally. The proposed learning strategy allows to enlarge inter-class distances while maintaining the intra-class structure of keystroke dynamics. We analyze the resulting representation space using the mathematical problem known as Circle Packing, which provides neighbourhood structures with a theoretical maximum inter-class distance. We finally prove experimentally the effectiveness of the proposed approach on a challenging task: keystroke biometric identification over a large set of 78,000 subjects. Our method achieves state-of-the-art accuracy on a comparison performed with the best existing approaches.
\end{abstract}


\end{frontmatter}


\section{Introduction}\label{sec:introduction}

In a global society migrating from physical services to digital platforms, identity management becomes critical. However, traditional physical user authentication cannot be directly applied in digital services. Keystroke biometric recognition enables the identification of users based on their typing behavior. Keystroke biometric systems are commonly placed into two categories: \textit{fixed-text}, characterized by a prefixed keystroke sequence typed by the user (e.g. passwords), and \textit{free-text}, characterized by arbitrary keystroke sequences (e.g. emails or transcriptions). Free-text systems must therefore consider different text content between training and testing, including typing errors. 

Keystroke dynamics authentication literature has been predominantly focused on verification tasks in fixed-text scenarios. Approaches based on statistical models (e.g. Hidden Markov Models) \cite{Ali}, Manhattan distances \cite{Vinnie1}, sample alignment (e.g. Dynamic Time Warping) \cite{2016_IEEEAccess_KBOC_Aythami}, and digraphs \cite{Bergadano} have achieved competitive results in fixed-text verification \cite{morales2015keystroke}. The performance in free-text scenarios remained far from those reached in the fixed-text verification approaches during the last decade. Partially Observable Hidden Markov Models were employed in \cite{Monaco} for free-text keystroke verification obtaining a competitive accuracy. More recently, the availability of large scale databases with millions of keystroke samples has allowed training deep models with very competitive performances in free-text scenarios \cite{TypeNet}. The architecture proposed in \cite{TypeNet}, called TypeNet, was trained using a Contrastive Loss function with performances six times better than previous approaches based on traditional statistical methods \cite{Monaco,Ceker}. Our purpose in the present paper is to improve further the state-of-the-art results of deep keystroke biometrics by introducing a new loss function expected to be also useful in other challenging recognition problems.

There are two main research lines to define such loss functions: \textit{i)} approaches based on Distance Metric Learning (DML) such as Contrastive Loss \cite{ContrastiveLoss}, Triplet Loss \cite{TripletLoss}, and their variants \cite{NPairLoss, chen2017beyond}; and \textit{ii)} with multi-class classifiers based on Softmax Loss functions and its variants \cite{CenterLoss, liu2016large, ArcFace}. Both research lines present advantages and disadvantages. 

In the first line of work (DML), the core idea is to train a function that maps input data into a new feature space where simple distances can serve to analyze and exploit the ``semantic'' structure of the input space \cite{ContrastiveLoss}. A DML approach serves to define a neighbourhood structure in the feature space based on a relationship between intra-class (between samples from the same class) and inter-class distances (between samples from different classes). In an ideal feature space, samples from the same class will remain ``near'' and samples from different classes will be pushed ``far''. Near and far can be defined based on simple distances like Euclidean. Noteworthy, most of the DML approaches in the literature are based on learning processes based on pairs of samples \cite{ContrastiveLoss, TripletLoss, CenterLoss}.

In the second line of work (i.e., using multi-class classifiers), there are some limitations stemmed from using classifiers. A classification algorithm is mostly associated to categorization tasks, but it can be used to tackle other representation learning problems. One example is the use of classification algorithms as feature extractors where models are trained for classification, and the outputs (usually the last layers of a deep network) are employed as features for other tasks \cite{ArcFace,Snoek15,Qi15}. However, using classification algorithms to learn discriminatory feature spaces exhibits limitations. On the one hand, the feature space learned might not be suitable for classes not seen during learning. On the other hand, the error propagated during learning is based on a scalar prediction (i.e., a label), which is a simplification of the whole problem at hand defined by intra- and inter-class neighborhood structures \cite{ContrastiveLoss}. To address these problems, some authors have proposed methods based on the joint supervision of Softmax Loss and DML to improve the discrimination power of the feature learned spaces \cite{CenterLoss, ArcFace}.

In the present work, we propose a novel DML approach (SetMargin Loss, SM-L) specifically designed to address the challenges associated to free-text keystroke identification where classes used in learning and inference are disjoint. The final aim is to identify the membership of the input data to a class unseen during learning. SM-L extends traditional DML approaches with a learning process guided by pairs of sets, which allows to enlarge inter-class distances while maintaining the intra-class structure. 

We will analyze the learned feature space generated with SM-L in comparison with other popular loss functions using the mathematical problem known as Circle Packing. The solution to the Circle Packing problem is a neighbourhood structure that guarantees a theoretical maximum inter-class distance. We propose using this method to gain understanding in the feature spaces obtained by DML approaches. Finally, we will prove experimentally the effectiveness of our proposed SM-L on a challenging task: keystroke biometric identification \cite{TypeNet}. The proposed approach outperforms other popular loss functions in this problem. 

In summary, the contributions of this work are: 
\begin{itemize}
    \item A new keystroke identification\textcolor{Black}{/verification} method based on a novel loss function called SetMargin Loss (SM-L).
    \item We introduce the Circle Packing problem as a novel way to gain insights into learned feature spaces.
    \item We experiment with the proposed SM-L on a challenging open-set keystroke biometric identification\textcolor{Black}{/verification} problem over 78,000 subjects, achieving state-of-the-art performance superior to related methods.
\end{itemize}

The rest of the paper is organized as follows. Next section summarizes the most popular DML approaches and presents the Circle Packing Problem as a way to analyze feature spaces. The third section describes the proposed SetMargin Loss and the fourth section presents the experiments and results. Finally, the work finish with the conclusions.

\section{Distance Metric Learning: Loss Functions}

The objective of metric learning is to generate distances $d$ between input data pairs either from the same or different classes (positive and negative pairs, respectively) useful for a certain task where a component of the distance is based on a learned model related to the task at hand. These distances $d$ can be defined, e.g., as Euclidean distances: 

\begin{equation}
\label{distance}
     d(\textbf{x}^{i},\textbf{x}^{j})= \left \| \textbf{f}(\textbf{x}^{i}|\textbf{w}) - \textbf{f}(\textbf{x}^{j}|\textbf{w})\right \|
\end{equation}

\noindent where $\textbf{w}$ are the weights of a model (typically a neural network), and $\textbf{f}(\textbf{x}^{i}|\textbf{w})$, $\textbf{f}(\textbf{x}^{j}|\textbf{w})$ are the model outputs (embedding vectors) for the inputs $\textbf{x}^{i}$ and $\textbf{x}^{j}$, respectively.

There are several metric learning approaches in the literature \cite{ContrastiveLoss, TripletLoss,  NPairLoss, CenterLoss}. Among these approaches, the \textit{Contrastive Loss} function \cite{ContrastiveLoss} is a popular example of DML technique with a history of success in many applications \cite{ TypeNet, Taigman, deb2019actions}. Let $\textbf{x}^{i}$ and $\textbf{x}^{j}$ each be a sample that together form a pair which is provided as input to a Siamese Neural Network \cite{Taigman} with shared weights $\textbf{w}$. The loss function $\mathcal{L}_{CL}$ is defined as follows:
\begin{equation}
\label{CL_loss}
     \mathcal{L}_{CL}= (1-L_{ij})\frac{d^2(\textbf{x}^{i},\textbf{x}^{j})}{2}+L_{ij}\frac{\max^2\left \{0, \alpha-d(\textbf{x}^{i},\textbf{x}^{j})\right \} }{2}
\end{equation}

\noindent where $L_{ij}$ is a label associated with each pair that is set to $0$ for positive pairs and $1$ for negative ones, and $\alpha \geq 0$ is a margin. As we can see, the Contrastive Loss learns intra- and inter-class distances in separate operations defined by $L_{ij}$.

The \textit{Triplet Loss} function \cite{TripletLoss, Facenet} appeared as a function to learn from positive and negative comparisons at the same time. A triplet is defined by three samples known as Anchor, Positive, and Negative. Anchor ($\textbf{x}^{i}_A$) and Positive ($\textbf{x}^{i}_P$) are samples from the same class $i$, while Negative ($\textbf{x}^{j}_N$) is a sample from a different class $j$. The Triplet loss function is defined as follows:

\begin{equation}
\label{TL_loss}
     \mathcal{L}_{TL}= \max \left \{0,d^2(\textbf{x}^{i}_A,\textbf{x}^{i}_P) - d^2(\textbf{x}^{i}_A,\textbf{x}^{j}_N) + \alpha \right \}
\end{equation}

\noindent where $\alpha$ is a margin between positive and negative pairs. In comparison with Contrastive Loss, Triplet Loss is capable of learning intra-class and inter-class structures in a unique operation (removing the label $L_{ij}$).

There are other metric learning methods designed to guide the learning process in different ways. The \textit{Center Loss} was proposed to minimize the intra-class distances of the deep features \cite{CenterLoss}. Center Loss combines the traditional Softmax with a loss function aimed to reduce the distance of feature vectors to an average feature vector calculated for each class (i.e., centroid). Similarly, the \textit{Magnet Loss} \cite{MagnetLoss16}  introduces a learning approach inspired in clustering techniques where the loss function depends on cluster optimization instead of traditional sample classification. \textit{N-Pair Loss} is an extension of Triplet Loss to several negative samples \cite{NPairLoss}, where triplets are conformed using one positive sample and $N$ multiple negative examples. More recently, researchers have proposed the \textit{Angular Softmax Loss} \cite{ArcFace} to improve face recognition performance. These methods improve the traditional Softmax Loss function by incorporating an angular learning objective.

\subsection{Circle Packing and Learned Feature Spaces}

\begin{figure}[t]
\centering
\includegraphics[width=0.6\columnwidth]{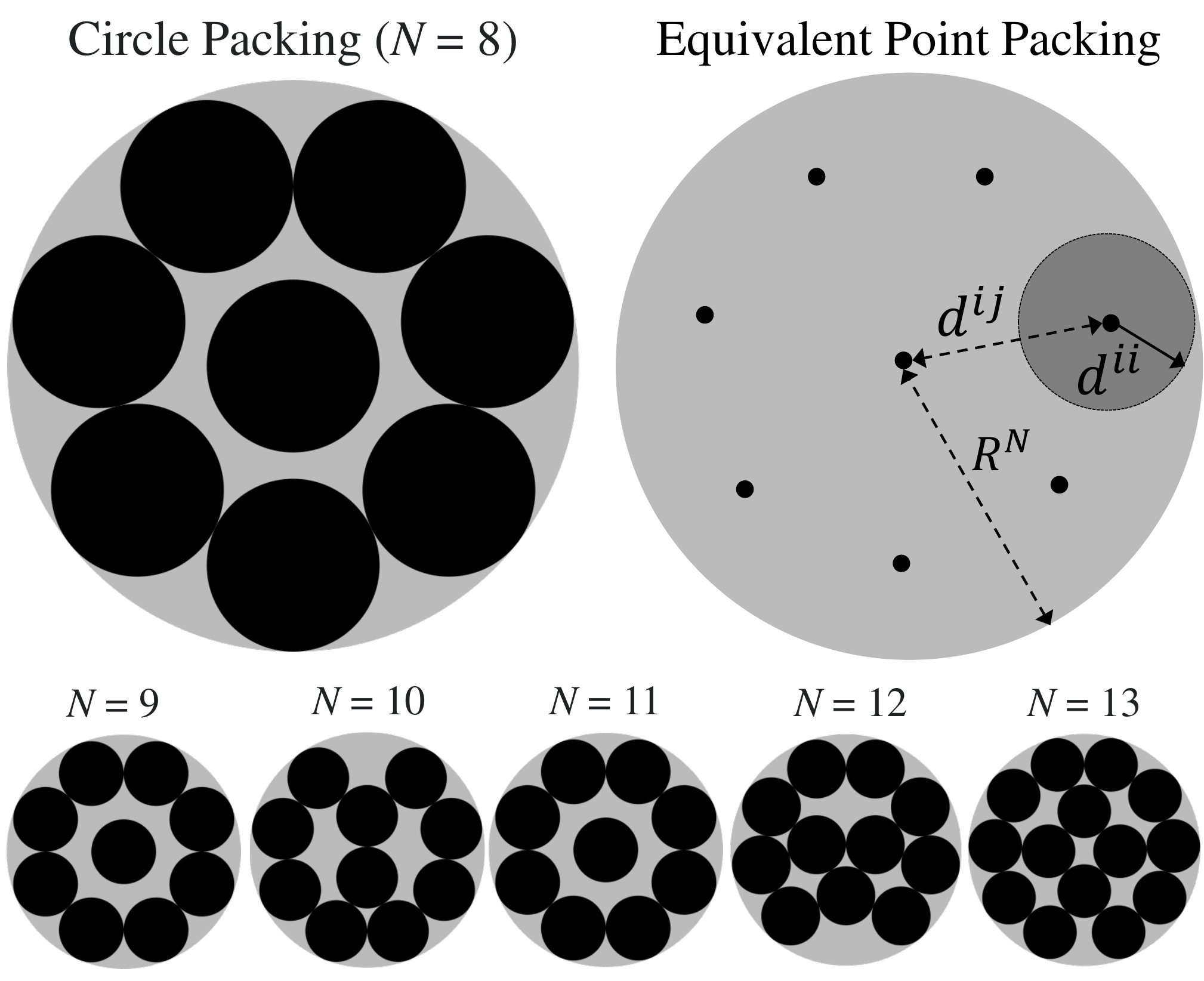} 
\caption{Circle Packing solution from $N=8$ to $N=13$ and equivalent Point Packing problem for $N=8$. $d^{ij}$ and $d^{ii}$ represent the inter- and intra-class distances respectively. \textcolor{Black}{$R^N$ is the minimum radius of the circular region $D$ that contains the unit circles.}}
\label{packing}
\end{figure}

In this section we introduce the Circle Packing problem to gain insights into the neighborhood structures learned in feature spaces. \textcolor{Black}{A packing of circles as defined in \cite{stephenson1990circle} is a collection (finite or infinite) of circles $P=\{C_1,...,C_N\}$ on a given (Riemann) surface $S$ with disjoint interiors (i.e., distinct circles in $P$ may be tangent, but cannot overlap). This is the general definition. In our specific case, the Riemann surface is a closed circular region $D \subset \mathbb{R}^2$ with the standard Euclidean metric, the packing $P$ is finite, and all circles $C_i$ have unit-radius. Thus, our objective can be defined as finding a Circle Packing $P$ such that the minimum distance between circles is maximized.} The result is a structure that minimizes the area outside the unit circles\textcolor{Black}{, and therefore the radius $R^N$ of the outer circle $D$}. The solution depends on the number of circles $N$ and the analytical demonstration varies depending on $N$ (i.e., there is not a unique way to solve the problem for different $N$). Figure \ref{packing} shows the solutions from $N=8$ to $N=13$ proved by \cite{Melissen94, Fodor00, Fodor03}. Note that, as also shown in Figure~\ref{packing}, the Circle Packing problem can be transformed into a Point Packing problem by replacing circles by their centers. This mathematical problem can be seen as an optimization task, and many computational approaches have been proposed to solve it \cite{Hifi07,Hifi09}. \textcolor{Black}{To the best of our knowledge, this is the first time that Circle Packing is used to analyze learned feature spaces.}


The solution to the Circle Packing problem maximizes the distances between circle centers $d^{ij}$ subject to fixed radii. Assuming that each circle is a class in our feature space, the distance $d^{ij}$ represents the inter-class distance (i.e., distance between samples from different classes) while $d^{ii}$ represents the intra-class distance (i.e., distance of samples from the same class). This formulation is specially useful in open-set classification problems, where we seek a feature space that: \textit{i)} maximizes inter-class distances; and \textit{ii)} minimizes intra-class distances. To achieve the maximum inter-class distance, it is necessary to distribute classes along all the space. \textcolor{Black}{It is important to keep in mind that in an open-set scenario the classes used during training are different to those used for testing.}       

We have conducted a toy example to visualize the feature space obtained by different loss functions and its similarity to the Circle Packing optimal solution. To this end, we use a subset of the ``Quick, Draw!" dataset \cite{QuickDraw}. This dataset comprises $50$ million drawings across $345$ different categories. This database is interesting because of the large intra-class variability in the different classes (e.g. there are hundreds of different ways to draw a plane). Each drawing is converted to a $28 \times 28$ grey scale image. In order to visualize the feature space learned by typical deep models, we train a Convolutional Neural Network (CNN) inspired in the popular VGG architecture \cite{VGG} and composed of: two Convolutional layers ($32$ and $64$ units, $3\times3$ filter bank size, ReLU activation), 2D Maxpooling layer, Dense layer (128 units, ReLU activation), Dense layer mapping the features into a 2D space (2 units, Linear activation), and Output layer (13 units, Softmax activation).

\begin{figure}[t]
\centering
\includegraphics[width=0.9\columnwidth]{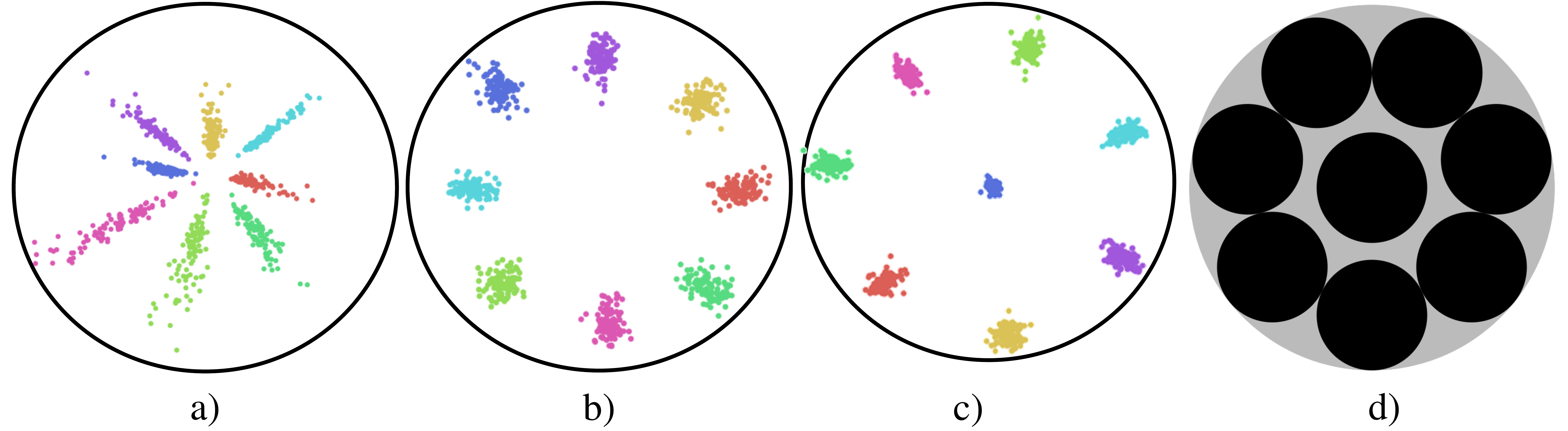} 
\caption{2D feature spaces ($N=8$) learned by: (a) Softmax Loss, (b) Contrastive Loss, (c) Triplet Loss, and (d) Circle Packing optimal solution for $N=8$.}
\label{loss_packing}
\end{figure}

We use $100$ images from the first $8$ classes of the ``Quick, Draw!" dataset to train a classifier (batch size = 32, Adam optimizer, learning rate = 0.01). Figure \ref{loss_packing} shows an example of how different loss functions define the feature space (for Contrastive Loss and Triplet Loss, we have removed the final output layer of the model). The feature spaces are generated plotting the output of the 2 units layer included in the CNN model. The feature space obtained by Softmax Loss has an intrinsic angular distribution as it is expected \cite{ArcFace}. The angular distribution obtained by Softmax is far from the Circle Packing solution, but it is not necessarily a wrong solution. Recent approaches based on softmax angular margin losses have achieved state-of-the-art performances in Face Recognition problems \cite{ArcFace}. Contrastive Loss \cite{ContrastiveLoss} maximizes the inter-class distances but fails to exploit all the available space. The feature space obtained by this loss function is similar to the one obtained by other loss functions that maximize inter-class distances in a joint supervision with Softmax \cite{CenterLoss}. Finally, Triplet Loss \cite{TripletLoss} shows a feature space that perfectly matches the optimal Circle Packing solution (see Figure \ref{loss_packing}). The feature space generated by Triplet Loss approximates the theoretical maximum inter-class distance in a feature space divided into $8$ circular regions.

\subsection{Limitations of the Circle Packing Solution}

Besides the several similarities between the optimal solution to the Circle Packing problem and the feature space generated with a specific learning strategy, we have to consider some limitations in this comparison: 

\begin{itemize}
    \item The results obtained by DML approaches are usually characterized by a separable Euclidean space. For approaches based on Euclidean spaces, the Circle Packing framework can be used to find \textcolor{Black}{a theoretical maximum} inter-class distance. However, defining each class region as a unit circle is not necessarily the best approach for all problems. The distance $d^{ii}$ can vary between classes (i.e., intra-class variability \textcolor{Black}{depends on the class) and we are projecting into a 2-dimensional feature space assuming low correlation between features}. In the present paper we simplify the problem assuming unit circles, but Circle Packing can be extended to circles with different area  \cite{stephenson2003circle} or ellipses with different shape \cite{birgin2013packing}.  
    \item Most of the learned feature spaces are characterized by more than $2$ dimensions. Extensions can be made to higher dimensions. \textcolor{Black}{In $3$ dimensions the equivalent problem is known as Sphere Packing; and Hypersphere Packing in higher dimensions.}
    \item In the present paper, we will show how these spaces are suitable for open-set classification problems. Nonetheless, a feature space defined by the Circle Packing solution is not necessarily the best solution for all machine learning problems, e.g., that solution does not guarantee good generalization properties to unseen data.   
\end{itemize}


\section{Proposed Method: SetMargin Loss (SM-L)}
\setlength{\belowdisplayskip}{0pt} \setlength{\belowdisplayshortskip}{0pt}
\setlength{\abovedisplayskip}{0pt} \setlength{\abovedisplayshortskip}{0pt}


The main challenges associated to free-text keystroke identification are: \textit{i)} large intra-class variability (i.e., the typing behavior of a given subject may vary occasionally); \textit{ii)} low inter-class variability (i.e., the typing behavior from different subjects might be similar); \textit{iii)} large number of classes (one per subject which can easily scale to several thousands); \textit{iv)} free-text scenario (i.e., identification must be performed with independence of the text typed); and  \textit{v)} small to moderate number of samples per class available to model the problem (i.e., 15 samples per class in our case). These challenges associated to keystroke biometric identification have made to fail recent machine learning approaches in this task for identifying a large number of subjects \cite{TypeNet}.

\begin{figure*}[t!]
\centering
\includegraphics[width=0.9\textwidth]{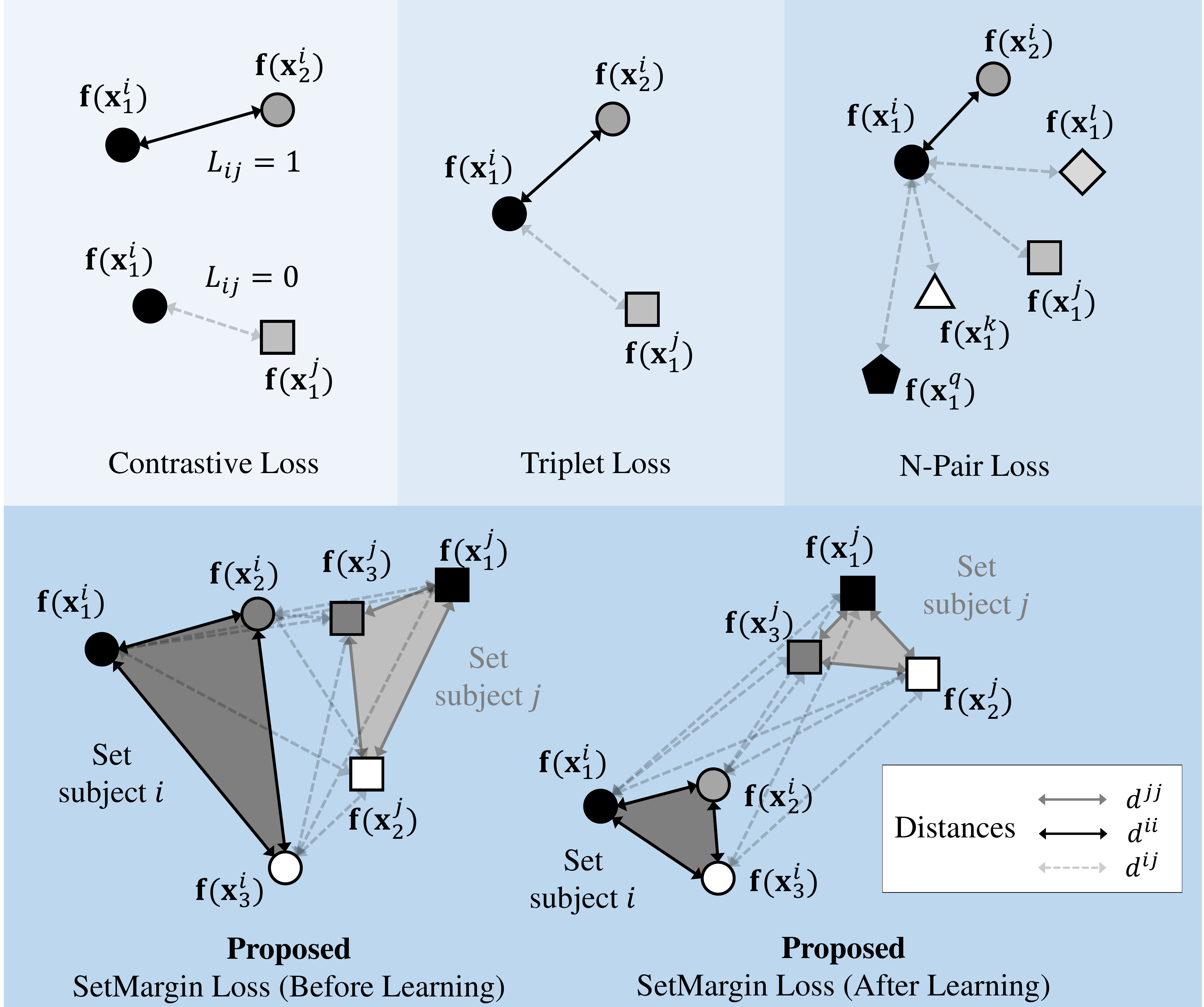}
\caption{Sets of distances considered in popular deep learning loss functions: Contrastive Loss, Triplet Loss, N-Pair Loss, and the proposed SetMargin Loss (SM-L) for a pair of sets with $G=3$ samples per set. Shapes represent different classes while color indicates different samples for the same class.}
\label{intuition}
\end{figure*}

Here we propose to overcome these challenges with an extension of Tripet Loss. In comparison with Contrastive Loss, Triplet Loss allows to model the relationship between positive and negative samples in a unique operation (see Figure \ref{intuition} and Eq. \ref{TL_loss}). Both methods were developed to learn from comparisons made with pairs of samples $d(\textbf{x}^{i},\textbf{x}^{j})$. A learning process guided by pairs of samples may not be adequate for the possibly complex intra-class relationships between samples of the same class. With our proposed SetMargin Loss (SM-L) we propose to extend this learning strategy to pairs of sets instead of pairs of samples. This learning strategy allows to capture better intra-class dependencies while enlarging the inter-class differences in the feature space (see Figure \ref{intuition}). 

In practice, there are different ways to transform a sample-pair based learning into a sample-set learning process. We propose to evaluate two different implementations of our idea of set distances: \textit{SetMargin Contrastive Loss (SM-CL)} and \textit{SetMargin Triplet Loss (SM-TL)}.   

Let $\{\textbf{x}_{k}^i\}_{k=1,...,G^i}$ and $\{\textbf{x}_{q}^j\}_{q=1,...,G^j}$ be a pair of sets provided as input to the model. The \textit{SetMargin Contrastive Loss (SM-CL)} proposed in this work is an extension of Eq.~\ref{CL_loss} defined as follows: 
\begin{multline}
\label{AG_loss_CL}
     \mathcal{L}_{SM-CL}= \sum\limits_{k=1}^{G^i} \sum\limits_{q=k+1}^{G^i} \frac{d^{2}(\textbf{x}_{k}^i,\textbf{x}_{q}^i)}{2} + \beta \sum\limits_{k=1}^{G^i} \sum\limits_{q=1}^{G^j} \frac{\max^{2} \left \{0, \alpha - d(\textbf{x}_{k}^i,\textbf{x}_{q}^j)\right \} }{2} 
\end{multline}
\noindent where $\alpha$ is a margin, $d(\cdot)$ is the Euclidean distance defined in Eq. \ref{distance} and  $\beta$ is a constant that serves to weight the intra-class and inter-class distances. In our experiments $G^i=G^j=G$ is the number of samples per class and $\beta=2G$ is proportional to the number of learning samples per class.

The \textit{SetMargin Triplet Loss (SM-TL)} proposed in this work is an extension of traditional Triplet Loss (see Eq.~\ref{TL_loss}) to learn from pairs of sets instead of pair of samples. This extension adds the context of the set to the learning process resulting in a large-margin representation capable of improving the distance between classes. The loss function is calculated as follows: 


\begin{multline}
\label{AG_loss_TL}
     \mathcal{L}_{SM-TL}= \sum\limits_{k=1}^{G^i} \sum\limits_{q=k+1}^{G^i} \sum\limits_{l=1}^{G^j} (\max \left \{0, d^2(\textbf{x}^{i}_k,\textbf{x}^{i}_q) - d^2(\textbf{x}^{i}_k,\textbf{x}^{j}_l) + \alpha \right \}  + \\
    + \max \left \{0, d^2(\textbf{x}^{j}_k,\textbf{x}^{j}_q) - d^2(\textbf{x}^{j}_k,\textbf{x}^{i}_l) + \alpha \right \} ) 
\end{multline}

Note that we assume $G^i=G^j=G$ but the method can be directly extended to problems where $G^i \neq G^j$. The margin $\alpha$ is equal to $1.5$ in all our experiments. \textcolor{Black}{The literature has shown that triplet selection can significantly improve the quality of the learned spaces \cite{schroff2015facenet, hermans2017defense}. Our approach does not include a direct selection of triplets. Nonetheless, the $max$ function included in Eq. \ref{AG_loss_TL} is used to reduce the impact of ``easy triplets" (i.e., $d^2(\textbf{x}^{i}_k,\textbf{x}^{j}_l) > d^2(\textbf{x}^{i}_k,\textbf{x}^{i}_q) + \alpha$).} 

\begin{figure*}[t]
\centering
\includegraphics[width=\textwidth]{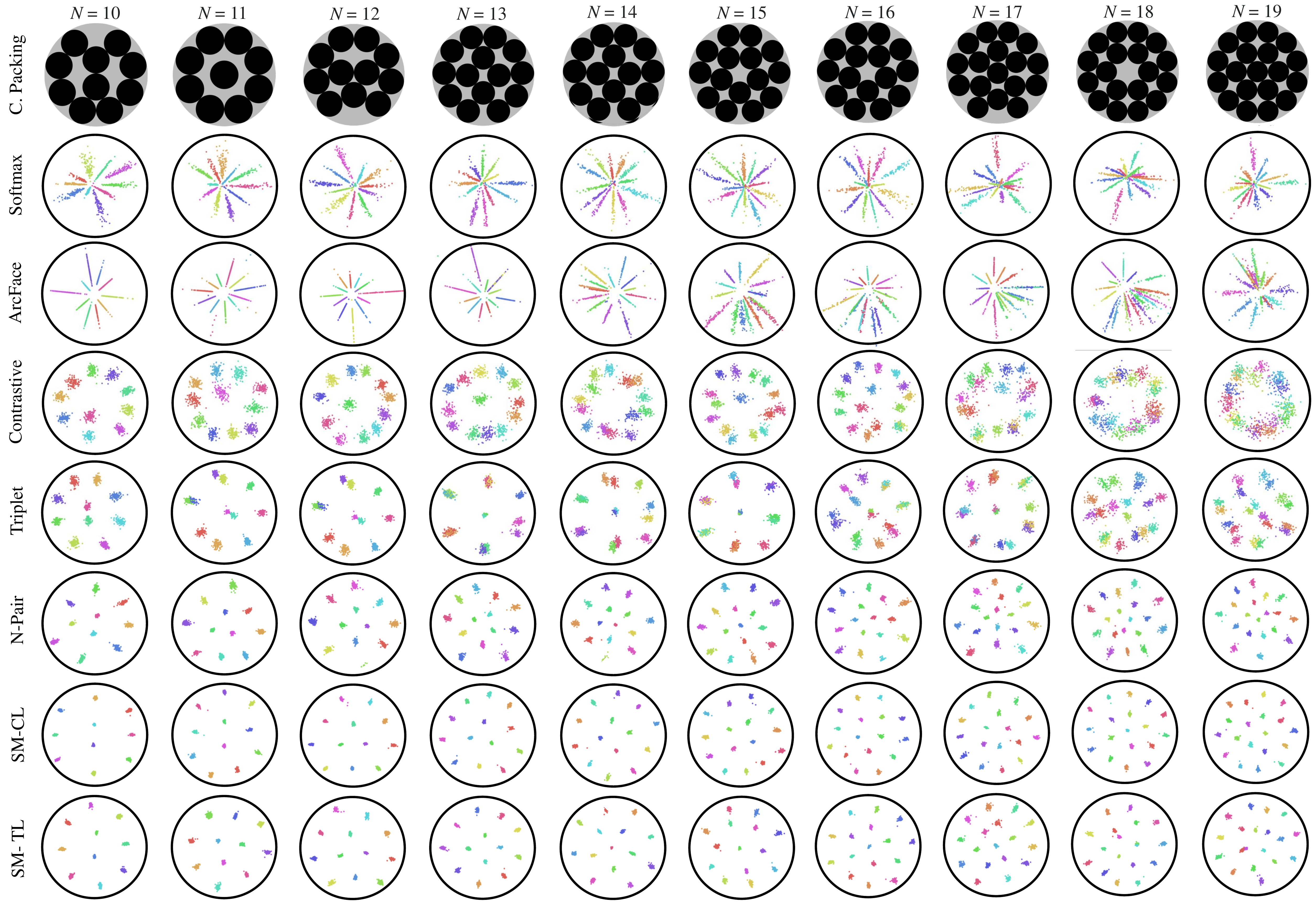} 
\caption{Circle Packing problem solutions from $N=10$ to $N=19$ and feature spaces learned by different loss functions. }
\label{loss_comparison}
\end{figure*}

\subsection{Intuition of the Learning Process} 

Without loss of generality let's assume a template composed of $3$ samples ($G=3$). $\mathcal{T}^i$ is the triangle whose vertices are each of the three embeddings of the template of subject $i$ and $d^{ij}$ is the distance between barycenters from $\mathcal{T}^i$ to $\mathcal{T}^j$. The \textit{SetMargin Loss} minimizes the areas of $\mathcal{T}^i$ and $\mathcal{T}^j$ while maximizing  the distance $d^{ij}$ (close to $\alpha$). Incorporating the template geometry into the learning objective we enrich the learned space generation process and this will result in better embedding representations (see Figure \ref{loss_comparison}).

The main characteristics of the proposed SM-L metric learning are: \textit{i)} it maximizes the inter-class distance while preserving intra-class compactness through batches composed by pairs of sets instead of pairs of samples; \textit{ii)} it produces highly discriminative feature spaces adequate for open-set classification problems where inference is conducted on samples of classes unseen in the learning process, and therefore the feature space is constructed over unseen class relationships; and \textit{iii)} it is able to learn highly discriminative feature spaces from a limited number of samples per class. The number of possible set combinations is very high and the set generation acts as a data augmentation technique.

\subsection{Comparison with other Loss Functions}

Figure \ref{loss_comparison} shows the feature space learned by different loss functions using the toy example presented in previous sections. The figure depicts the feature spaces for: Softmax Loss, ArcFace Loss \cite{ArcFace}, Contrastive Loss \cite{ContrastiveLoss}, Triplet Loss \cite{TripletLoss}, N-Pair Loss \cite{NPairLoss}, and our SetMargin Loss in the two proposed implementations: SM-CL and SM-TL. The angular distribution of classes observed for the Softmax Loss function is enhanced by the ArcFace Loss function \cite{ArcFace}. Contrastive Loss shows a feature space with a common pattern where all classes are distributed in the exterior regions and  one class is located at center. These spaces are similar to those obtained by other loss functions such as Center Loss \cite{CenterLoss}. For large number of classes, this type of distribution is highly inefficient. Triplet Loss tends to create spaces with structures similar to those of the Circle Packing solution, but fails with $N$ greater than $10$. N-Pair Loss is based on a learning process guided by inter-class comparisons. Thus, N-Pair Loss improves the margin between classes with respect to Triplet Loss. The proposed SetMargin Loss shows feature spaces very similar to those obtained by the optimal solution to the Circle Packing problem. The feature space obtained by our method guarantees a good separation between classes, close to the theoretical maximum. 

\textcolor{Black}{We propose two quantitative metrics to evaluate the distribution of classes in the learned spaces: minimum distance between centroids ($\delta$), and intra-cluster dispersion ($\rho$). The minimum distance between centroids $\delta$ is calculated as:}

\begin{equation}
\label{mind_delta}
     \textcolor{Black}{\delta = \frac{1}{N}  \sum\limits_{i=1}^{N} \min_j ||\textbf{c}^i-\textbf{c}^j||, (i\neq j)}
\end{equation}

\noindent \textcolor{Black}{where $N$ is the number of classes and $\{\textbf{c}^i,\textbf{c}^j\}$ are the centroids of the embeddings of the classes $i$ and $j$. The intra-cluster dispersion $\rho$ is calculated as:}

\begin{equation}
\label{mind_rho}
     \textcolor{Black}{\rho = \frac{1}{N} \sum\limits_{i=1}^{N} \left(\frac{1}{L^i} \sum\limits_{k=1}^{L^i} ||\textbf{c}^i-\textbf{f}(\textbf{x}^{i}_k)||\right)}
\end{equation}

\noindent \textcolor{Black}{where $L^i$ is the total number of data points (i.e., number of samples) of the class $i (L^i \geq G^i)$ and $\textbf{f}(\textbf{x}^{i}_k)$ is the embedding vector $k$ of the class $i$. The maximum $\delta$ for a given $N$ can be calculated using the Circle Packing solution as:}

\begin{equation}
\label{mind_delta_max}
     \textcolor{Black}{\delta^{N}_{max} = \frac{1}{2R^{N}-1}}
\end{equation}

\noindent \textcolor{Black}{where $R^{N}$ is the minimum radius of the circular region $D$ that contains the $N$ unit circles. In a similar way, the maximum distance between centers of adjacent unit circles is calculated as $\delta^{N}_{CP} = 1/2R^{N}$.}

\textcolor{Black}{Table \ref{tab:distances} presents the $\rho$ and $\delta$ obtained for learned spaces trained with Contrastive Loss (CL), Triplet Loss (TL), and our proposed implementations (SM-CL and SM-TL). Note that for all $N$, the proposed SM-CL and SM-TL outperform the previous implementations in both $\rho$ and $\delta$. Larger values of $\delta$ mean larger distance between classes (i.e., high inter-class distance), while lower values of $\rho$ mean lower distance between samples from the same class (i.e., low intra-class distance). Table \ref{tab:distances} also includes the theoretical maximum distance obtained by the optimal Circle Packing solution. The results show that the proposed distances (SM-CL and SM-TL) outperform the Circle Packing optimal solution $\delta^{N}_{CP}$ with distances close to the theoretical maximum $\delta^{N}_{max}$. This is possible as the embeddings of the proposed learned feature spaces tend to be clustered in the border of the unit circle instead of the center.}

\begin{table}[t]
\small
\renewcommand{\arraystretch}{1.0}
  \begin{center}
  \caption{\textcolor{Black}{Results of the minimum distance between centroids $\delta$ and intra-cluster dispersion $\rho$ (in parenthesis, multiplied by $100$) for the different loss functions. We also include the values of $\delta$ for the Circle Packing optimal solution ($\delta_{CP}$) and the theoretical maximum ($\delta_{max}$).}}
  \label{tab:distances}
\smallskip
    \begin{tabular}{l|cccccc} 
      \hline
      \textcolor{Black}{\textbf{Method}} & \textcolor{Black}{$N$=12} & \textcolor{Black}{$N$=14} & \textcolor{Black}{$N$=16} & \textcolor{Black}{$N$=18} & \textcolor{Black}{$N$=20}\\
     \hline
       \textcolor{Black}{$\delta_{CP}$}  & \textcolor{Black}{$0.25$} & \textcolor{Black}{$0.23$} & \textcolor{Black}{$0.22$} & \textcolor{Black}{$0.21$} & \textcolor{Black}{$0.19$}\\ 
       \textcolor{Black}{$\delta_{max}$}  & \textcolor{Black}{$0.33$} & \textcolor{Black}{$0.30$} & \textcolor{Black}{$0.28$} & \textcolor{Black}{$0.26$} & \textcolor{Black}{$0.24$}\\ 
      \hline
      \textcolor{Black}{Contrastive \cite{ContrastiveLoss}}  & \textcolor{Black}{$0.23(0.55)$} & \textcolor{Black}{$0.15(0.29)$} & \textcolor{Black}{$0.10(1.09)$} & \textcolor{Black}{$0.09(1.25)$} & \textcolor{Black}{$0.08(1.26)$}\\ 
      \textcolor{Black}{Triplet \cite{TripletLoss}}  & \textcolor{Black}{$0.22(0.27)$} & \textcolor{Black}{$0.15(0.30)$} & \textcolor{Black}{$0.09(0.34)$} & \textcolor{Black}{$0.08(0.39)$} & \textcolor{Black}{$0.08(0.39)$}\\ 
      \hline
      \textcolor{Black}{SM-CL [ours]}  & \textcolor{Black}{$0.26(0.13)$} & \textcolor{Black}{$0.25(0.14)$} & \textcolor{Black}{$0.25(0.20)$} & \textcolor{Black}{$0.21(0.18)$} & \textcolor{Black}{$0.20(0.21)$}\\ 
      \textcolor{Black}{SM-TL [ours]}  & \textcolor{Black}{$0.30(0.15)$} & \textcolor{Black}{$0.26(0.18)$} & \textcolor{Black}{$0.25(0.19)$} & \textcolor{Black}{$0.21(0.20)$} & \textcolor{Black}{$0.20(0.21)$}\\ 
    \end{tabular}
  \end{center}
   
\end{table}

\section{Experiments}

The SetMargin Loss has been developed mainly to improve the learned space in open-set classification scenarios (anticipating that it will be also very helpful in many other machine learning scenarios). To evaluate our loss function, we therefore propose to experiment on a challenging \textit{keystroke biometric identification} task, where subjects are identified based on their typing behavior \cite{2016_IEEEAccess_KBOC_Aythami, TypeNet, monrose1997authentication, gunetti2005keystroke, banerjee2012biometric}.

\subsection{Dataset}
\label{aalto}
Our experiments are conducted with the Aalto University Dataset \cite{Dhakal} that comprises keystroke sequences from $168$,$000$ subjects. Specifically, we employ the first $78$,$000$ subjects available in the database. The acquisition task asked subjects to memorize English sentences and then to type them as quickly and accurately as they could. The English sentences were selected randomly from a set of $1$,$525$ examples taken from the Enron mobile email and Gigaword Newswire corpus. The example sentences contained a minimum of $3$ words and a maximum of $70$ characters. Note that the sentences typed by the subjects could contain a bit more than $70$ characters because each subject could forget or add new characters when typing. All subjects in the database completed $15$ sessions with a different sentence in each session on either a desktop or laptop keyboard. See \cite{Dhakal} for more details including demographic and acquisition information. 

\begin{figure*}[t!]
\centering
\includegraphics[width=\textwidth]{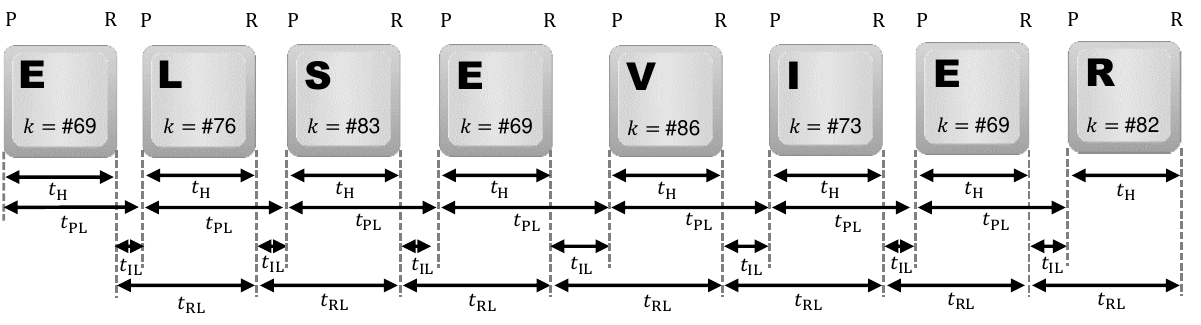}
\caption{Example of the 37 temporal features extracted from the keystroke sequence ``ELSEVIER'': $8$ $\times$ Hold time ($t_{\textrm{H}}$) + $7$ $\times$ Inter-key Latency ($t_{\textrm{IL}}$), $7$ $\times$ Press Latency ($t_{\textrm{PL}}$), $7$ $\times$ Release Latency ($t_{\textrm{RL}}$), $8$ $\times$ key codes ($k$). P = key Press event; R = key Release event.}
\label{features_fig}
\end{figure*}

\subsection{Pre-processing and Keystroke Dynamics} \label{features}

The keystroke raw data comprises a three dimensional time series including (see Figure \ref{features_fig}):  key press timestamps, key release timestamps, and the keycodes. Timestamps are in UTC format with millisecond resolution, and the keycodes are integers between $0$ and $255$ according to the ASCII code. The input to the model comprises sequences of keycodes plus $4$ temporal features: \textit{(i)} Hold Latency: the elapsed time between press and release key events; \textit{(ii)} Inter-key Latency: the elapsed time between releasing a key and pressing the next key; \textit{(iii)} Release Latency: the elapsed time between two consecutive release events; and \textit{iv)} Press Latency: the elapsed time between two consecutive press events. These $4$ features are commonly used in both fixed-text and free-text keystroke biometric systems \cite{Alsultan}. This feature extraction process results in a \textcolor{Black}{$K \times 5$} feature vector where textcolor{Blue}{$K$} is the number of keys pressed (note that keycode is added for each key pressed). In order to train the model with sequences of different lengths \textcolor{Black}{$K$} within a single batch, we truncate the end of the input sequence when \textcolor{Black}{$K>M$} and zero pad at the end when \textcolor{Black}{$K<M$}, in both cases to the fixed size $M$. The size of the time dimension $M$ was fixed to $M=50$, which was determined heuristically based on the characteristics of the dataset used in the experiments (see Sect.~\ref{aalto}). Error gradients are not computed for the padded zeros which do not contribute to the loss function.

\subsection{Implementation Details: RNN Model \textcolor{Black}{ and Experimental Protocol}}
\label{model}

In our experiments we used the architecture proposed in \cite{TypeNet}: TypeNet. TypeNet is a RNN architecture composed of two Long Short-Term Memory (LSTM) layers of $128$ units, and an initial Masking layer. Between the LSTM layers, the model performs batch normalization and dropout at a rate of $0.5$ to avoid overfitting. Additionally, each LSTM layer has a dropout rate of $0.2$. The output of the model $\textbf{f}($\textbf{x}$)$ is an array of size $128$ that we use as an embedding feature vector.

In our experiments, a batch was composed of $256$ set pairs ($\{\textbf{x}_{k}^i\}_{k=1,2,3}$, $\{\textbf{x}_{q}^j\}_{q=1,2,3}$) randomly chosen from the dataset available for learning. The number of possible set pairs is at billions scale. We used $500$ batches per epoch. The learning converges with less than $40$ epochs which means around $5$M set pairs in total.


\textcolor{Black}{\textbf{Training protocol:}} The RNN model was trained using the first $68$,$000$ subjects in the dataset according to the method proposed in \cite{TypeNet}. From the remaining $100$,$000$ subjects, we employed another $10$,$000$ subjects to perform the evaluation of the different loss functions, so there is no data overlap between the two groups of subjects. The distance between two keystroke sequences was computed by averaging the Euclidean distances between the $T$ gallery embedding vectors $\textbf{f}(\textbf{x}_{g}^{i})$ and the query embedding vector $\textbf{f}(\textbf{x}_{q}^{j})$ as follows:
\begin{equation}
\label{score}
     d_{i,j}= \frac{1}{T}\sum_{g=1}^{T}||\textbf{f}(\textbf{x}_{g}^{i})-\textbf{f}(\textbf{x}_{q}^{j})||
\end{equation}

\textcolor{Black}{The experiments include two scenarios: identification and verification}. The results reported in the next section are computed in terms of Rank-$n$ \textcolor{Black}{and Equal Error Rate (EER). We used the same trained models for both scenarios.}

\textcolor{Black}{\textbf{Identification protocol:} As a $1$:$N$ problem, the identification accuracy varies depending on the size of the background set. The background is conformed with identities (i.e., classes) not used in the learning process (i.e., open-set problem). The goal is to identify the query identity among all the background subjects. In our experiments, the size of the background was equal to $5$,$000$ identities (i.e., subjects). Additionally, our experiments include another $5$,$000$ subjects employed as query set for a total of $10$,$000$ different subjects in testing ($5$K background + $5$K query). We divided the $15$ keystroke sequences available for each subject into a gallery set (the first $10$ keystroke sequences) and a query set (the remaining $5$ keystroke sequences). Remember that it is a free-text scenario, so the model must identify the subject typing different keystroke sequences between gallery and query. We evaluated the identification accuracy by averaging the distance between the query set of samples $\textbf{x}_{q}^{j}$, with $q=1,...,5$ belonging to the subject $j$ and the gallery set $\textbf{x}_{g}^{i}$, with $g=1,...,10$  belonging to all $5$,$000$ background subjects $i$. Rank-$n$ is a measure of $1$:$N$ identification system performance. A Rank-$1$ means that $d_{i,J}<d_{I,J}$ for any $i \neq I$.} We then identify a query subject $J$ to be present in the gallery as subject $I$ as follows \cite{morales2020keystroke}:

\begin{equation}
\label{mindistance}
     I = \arg\min_i d_{i,J}
\end{equation}

\textcolor{Black}{This identification protocol did not include a decision threshold. The identity with the minimum distance was chosen among all the background subjects.}

\begin{figure}[t!]
\centering
\includegraphics[width=0.4\columnwidth]{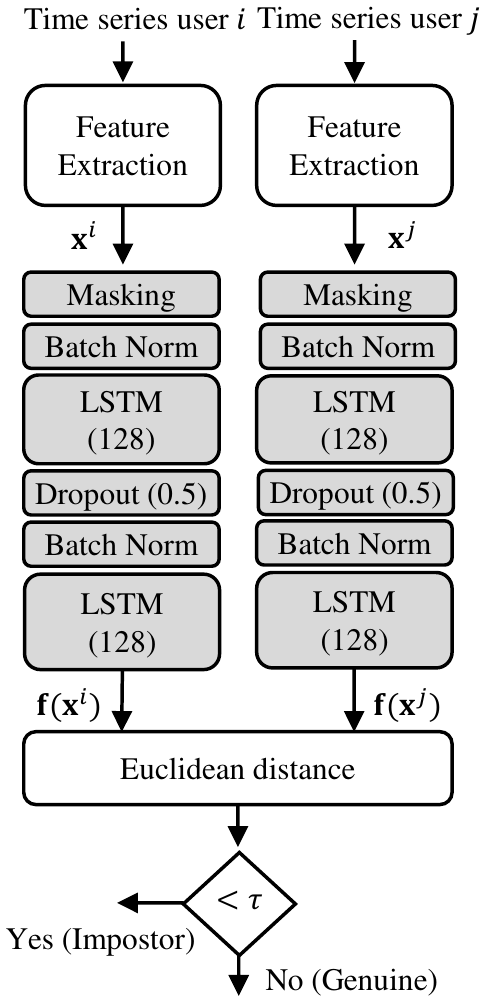}
\caption{\textcolor{Black}{Block diagram of the keystroke biometric verification system ($1:1$ comparison) based on the TypeNet \cite{TypeNet} architecture. $\tau$ is a decision threshold.}}
\label{architecture}
\end{figure}

\textcolor{Black}{\textbf{Verification protocol:} The EER is a measure of $1$:$1$ verification system performance. EER is defined as the operational point where False Rejection and False Acceptance are equal. The goal is to verify the identity of a query sample using its corresponding gallery set. This experiment includes a total of $5$,$000$ different subjects in testing. We divided the $15$ keystroke sequences available for each subject into a gallery set (the first $5$ keystroke sequences) and a query set (the last $5$ keystroke sequences). The score $d_{i,j}$ was obtained as the average distance between the query vector and the set of gallery samples. In this experiment, each query sample was evaluated separately for a total number of $50$,$000$ genuine scores ($5$ query samples $\times$ $5$,$000$ subjects). The impostor scores were obtained choosing one query sample per subject for a total number of $24$,$995$ scores ($5$,$000$ $\times$ $4$,$999$). We evaluated the verification accuracy averaging the EER obtained for each subject. Figure \ref{architecture} presents the block diagram of the verification protocol for a comparison $1:1$.}

\subsection{Results}
\label{experiments_results}

Table \ref{tab:table_acc} presents the identification performance of TypeNet~\cite{TypeNet} incorporating our proposed SetMargin Loss (\textit{SM-CL} and \textit{SM-TL}) in comparison to other popular loss functions: Triplet Loss \cite{TripletLoss}, Contrastive Loss \cite{ContrastiveLoss}, Softmax, Deep Linear Discriminant Analysis DeepLDA \cite{wu2017deep}, Quadruplet Loss \cite{chen2017beyond}, and N-Pair Loss \cite{NPairLoss}. We also include there as reference the results of two competitive algorithms for free-text keystroke biometrics based on statistical models: Partially Observable Hidden Markov Models \cite{Monaco}, and Digraphs-SVM \cite{Ceker}. Note that all approaches were trained using the same number of training sequences. Depending on the ML approach, the number of samples employed to compute the loss function varies. In order to make a fair comparison between loss functions, the batch size of the different approaches has been modified to incorporate the same number of samples per batch ($1$,$000$ samples per batch).

\begin{table}[t]
\small
\renewcommand{\arraystretch}{1.2}
  \begin{center}
   \caption{\textcolor{Black}{Identification performance in terms of} \textcolor{Black}{Rank-$n$} accuracy for different methods in the literature. $G$ is the number of samples conforming each set of samples employed to train the SetMargin Loss. \textcolor{Black}{Background dataset of $5$,$000$ subjects.}}
   \label{tab:table_acc}
\smallskip
    \begin{tabular}{l|ccc} 
      \hline
      \textbf{Method} & \textbf{Rank-1} & \textcolor{Black}{\textbf{Rank-5}} & \textcolor{Black}{\textbf{Rank-20}}\\
     \hline
      Digraph \cite{Ceker} & $0.5\%$ & \textcolor{Black}{$0.9\%$} & \textcolor{Black}{$1.2\%$}\\ 
      POHMM \cite{Monaco} & $6.1\%$ & \textcolor{Black}{$10.3\%$} & \textcolor{Black}{$13.8\%$}\\
      \hline
      TypeNet: \textit{Contrastive Loss} \cite{TypeNet}  & $17.8\%$  & \textcolor{Black}{$31.5\%$} & \textcolor{Black}{$38.9\%$}\\
      TypeNet: \textit{DeepLDA} \cite{wu2017deep}  & $34.2\%$ & \textcolor{Black}{$63.2\%$} & \textcolor{Black}{$84.2\%$}  \\
      TypeNet: \textit{Softmax} & $37.9\%$ & \textcolor{Black}{$64.9\%$} & \textcolor{Black}{$84.4\%$}\\
      TypeNet: \textit{Triplet Loss} \cite{TripletLoss} & $38.2\%$ & \textcolor{Black}{$68.2\%$} & \textcolor{Black}{$88.5\%$}\\
      TypeNet: \textit{Quadruplet Loss} \cite{chen2017beyond} & $38.6\%$  & \textcolor{Black}{$68.7\%$} & \textcolor{Black}{$87.9\%$}\\
      TypeNet: \textit{N-Pair Loss} \cite{NPairLoss} & $38.7\%$ & \textcolor{Black}{$67.7\%$} & \textcolor{Black}{$87.0\%$} \\
      \hline
      TypeNet: \textit{SM-CL, G=3} & $31.0\%$ & \textcolor{Black}{$59.9\%$} & \textcolor{Black}{$82.7\%$}\\
      TypeNet: \textit{SM-CL, G=6} & $37.5\%$ & \textcolor{Black}{$67.0\%$} & \textcolor{Black}{$86.8\%$}\\
      TypeNet: \textit{SM-CL, G=9} & $36.7\%$ & \textcolor{Black}{$65.8\%$} & \textcolor{Black}{$86.3\%$}\\
      TypeNet: \textit{SM-TL, G=3} & $39.4\%$ & \textcolor{Black}{$68.3\%$} & \textcolor{Black}{$88.1\%$}\\
      TypeNet: \textit{SM-TL, G=6} & $45.8\%$ & \textcolor{Black}{$73.9\%$} & \textcolor{Black}{$91.0\%$}\\
      TypeNet: \textit{SM-TL, G=9} & $45.3\%$ & \textcolor{Black}{$72.4\%$} & \textcolor{Black}{$89.5\%$}\\
      \hline
      
    \end{tabular}
  \end{center}
   
\end{table}

The results show how Triplet Loss obtains an accuracy two times higher than Contrastive Loss and similar performance than Softmax, Quadruplet Loss \cite{chen2017beyond}, and N-Pair Loss \cite{NPairLoss}. In comparison with traditional statistical approaches \cite{Monaco,Ceker}, the Deep Learning model (TypeNet) is clearly superior. The proposed SM-CL obtains a much higher performance than traditional Contrastive Loss but the accuracy achieved is still under other loss functions. Finally, SM-TL improves the best related loss function (N-Pair Loss) by $18\%$ relatively with a Rank-1 accuracy of $45.8\%$. SM-TL is able to capture the intra-class structure of samples from the same class and at the same time maximizes the inter-class distance. This learning process is appropriate for open-set classification tasks where query samples are matched to multiple different classes not seen during learning ($5$,$000$ in our experiments). \textcolor{Black}{These accuracies increase up to $73.9\%$ and $91\%$ for Rank-5 and Rank-20 respectively.} 

The results prove how the performance improves when incorporating sets of samples into the loss function. The superior performance of SM-TL cannot be attributed exclusively to the larger number of samples included in the computation of the loss function. As an example SM-CL showed lower performance with the same number of samples and N-Pair loss showed lower performance with larger number of samples. We have not included the performance of ArcFace in our comparison because of the poor results obtained. This poor performance can be caused by different factors including the low number of samples available per subject (only $15$ in contrast with hundreds of samples in \cite{ArcFace}), the architecture (RNN instead of CNN), or parameter tuning. 

\textcolor{Black}{Table \ref{tab:table_eer} presents the verification performance of our proposed SetMargin Loss (\textit{SM-CL} and \textit{SM-TL}) and other popular methods and loss functions. The verification scenario is characterized by higher accuracies in comparison with the identification experiments. In this case, the proposed method (SM-TL) is capable of achieving a performance of $1.85\%$ of EER. The method shows superior performance than previous approaches and popular loss functions. Nonetheless, in the verification scenario the margin of improvement is lower than in the identification experiment. It should be noted that we used the same model for both the identification and verification scenarios. The results demonstrate the high discrimination capacity of the learned spaces for both identification and verification.}

\begin{table}[t]
 
\small
\renewcommand{\arraystretch}{1.2}
  \begin{center}
  \caption{\textcolor{Black}{Verification performance in terms of Equal Error Rate (EER) for different methods in the literature. $G$ is the number of samples conforming each set of samples employed to train the SetMargin Loss. The best performance is obtained for $G=6$.}}
  \label{tab:table_eer}
\smallskip
    \begin{tabular}{l|c} 
      \hline
      \textcolor{Black}{\textbf{Method}} & \textcolor{Black}{\textbf{EER}}\\
     \hline
      \textcolor{Black}{Digraph \cite{Ceker}} & \textcolor{Black}{$43.1\%$} \\ 
      \textcolor{Black}{POHMM \cite{Monaco}} & \textcolor{Black}{$24.7\%$} \\
      \hline
      \textcolor{Black}{TypeNet: \textit{Contrastive Loss} \cite{TypeNet}}  & \textcolor{Black}{$5.40\%$}   \\
      \textcolor{Black}{TypeNet: \textit{DeepLDA} \cite{wu2017deep}}  &  \textcolor{Black}{$4.21\%$}  \\
      \textcolor{Black}{TypeNet: \textit{Softmax}} &  \textcolor{Black}{$10.8\%$}  \\
      \textcolor{Black}{TypeNet: \textit{Triplet Loss} \cite{TripletLoss}} &  \textcolor{Black}{$2.20\%$}  \\
      \textcolor{Black}{TypeNet: \textit{Quadruplet Loss} \cite{chen2017beyond}} &  \textcolor{Black}{$2.33\%$}  \\
      \textcolor{Black}{TypeNet: \textit{N-Pair Loss} \cite{NPairLoss}} &  \textcolor{Black}{$2.51\%$}  \\
       \textcolor{Black}{TypeNet: \textit{SM-CL, G=6}} &  \textcolor{Black}{$2.42\%$}  \\
       \textcolor{Black}{TypeNet: \textit{SM-TL, G=6}} &  \textcolor{Black}{$1.85\%$}  \\
      \hline
      
    \end{tabular}
  \end{center}
  
\end{table}

\subsubsection{Computational Load}

Metric Learning approaches suffer from data expansion when batches are conformed by pairs or triplets of samples instead of individual samples. This expansion offers some advantages (e.g. data augmentation), but also increases the computational load. The proposed SM-L learning process is defined by pairs of sets, instead of pairs/triplets of samples. Thus, our method exponentially increase the number of possible combinations. However, the convergence of the learning process is relatively fast and affordable for a personal computer with high specifications. All the experiments presented in this work were made with an Intel Core i7-8760H CPU @ 2.2Ghz, 32 GB RAM, NVIDIA GeForce RTX2080. As an example, the time needed to learn the SM-CL and SM-TL models used in our experiments were 7.6 and 8.8 hours, respectively.


\section{Conclusions}
\label{conclusions}
We have presented a new Distance Metric Learning approach called SetMargin Loss (SM-L). Our approach improves intra-class and inter-class structures in learned spaces, which is specially useful (among other machine learning problems) for open-set classification. We have also introduced the Circle Packing problem as a novel way to gain insights into the feature space of learned representations. A feature space that satisfies the Circle Packing problem guarantees a theoretical maximum inter-class distance given compact intra-class distances. Our experiments suggest that SM-L is capable of obtaining a feature space close to the Circle Packing optimal solution.

We have finally applied SM-L to keystroke biometric Identification using the Aalto University Dataset \cite{Dhakal}. Our experiments, conducted over a learning set with typing sequences from $68$,$000$ subjects and evaluated over a testing set with $10$,$000$ subjects, demonstrate the superior performance of the proposed approach over other popular loss functions. The proposed approach showed an accuracy (Rank-1 for identification and EER for verification) significantly superior than traditional statistical methods and $18\%$ better (relatively) than Softmax, Triplet, and N-Pair Losses. \textcolor{Black}{ The proposed SM-TL approach obtained a Rank-1 accuracy of $45.3\%$ and $1.85\%$ of EER. This performance is still far from the most accurate biometrics modalities, but it provides new opportunities in applications in the digital domain (e.g., online authentication, digital forensics). The EER under $2\%$ obtained for the user verification scenario demonstrates the potential of keystroke dynamics in large scale user authentication applications.}

\textcolor{Black}{For future work we suggest going deeper in the theory behind the proposed methods in order to seek theoretical limits on the achievable performance and to inspire new learning methods. Finally, we also plan to explore the developed methods in other problems beyond the ones explored here both in supervised and unsupervised learning.}

\section*{Acknowledgements}
This work has been supported by projects: PRIMA (MSCA-ITN-2019-860315), TRESPASS-ETN (MSCA-ITN-2019-860813), BIBECA (RTI2018-101248-B-I00 MINECO), edBB (UAM), and Instituto de Ingenieria del Conocimiento (IIC). A. Acien is supported by a FPI fellowship from the Spanish MINECO.
\bibliography{bibliography}

\end{document}